\documentclass{llncs}
\usepackage{makeidx}  % allows for indexgeneration
\usepackage{subfig}
\begin{document}
\mainmatter              % start of the contributions
\title{Hedging Algorithms and Repeated Matrix Games}
\titlerunning{Hedging and Matrix Games}  % abbreviated title 
\author{Bruno Bouzy \and Marc M\'etivier \and Damien Pellier}
\authorrunning{Bruno Bouzy et al.} % abbreviated author list
\institute{LIPADE, Universit\'e Paris Descartes, FRANCE,\\
\email{\{bruno.bouzy, marc.metivier, damien.pellier\}@parisdescartes.fr} 
}

\maketitle     

\begin{abstract}
Playing repeated matrix games (RMG) while maximizing the cumulative returns is a basic method to evaluate multi-agent learning (MAL) algorithms. Previous work has shown that $UCB$, $M3$, $S$ or $Exp3$ algorithms have good behaviours on average in RMG. Besides, hedging algorithms have been shown to be effective on prediction problems. An hedging algorithm is made up with a top-level algorithm and a set of basic algorithms. To make its decision, an hedging algorithm uses its top-level algorithm to choose a basic algorithm, and the chosen algorithm makes the decision. This paper experimentally shows that well-selected hedging algorithms are better on average than all previous MAL algorithms on the task of playing RMG against various players. $S$ is a very good top-level algorithm, and $UCB$ and $M3$ are very good basic algorithms. Furthermore, two-level hedging algorithms are more effective than one-level hedging algorithms, and three levels are not better than two levels. 
\end{abstract}

\section{ Introduction } \label{sectionIntro}

Multi-agent learning (MAL) questions how to learn when there are several learners in the environment. MAL is the heir of reinforcement learning, game theory, and multi-agent systems. Work in MAL can be divided into 5 agendas \cite{shoham-powers-grenager-aij-2007}: computational, descriptive, normative, prescriptive cooperative, and prescriptive non cooperative. The background of this paper is the prescriptive non cooperative agenda in which the goal of an agent is to maximize its cumulative return over time. In this context, many MAL algorithms already exist \cite{littman-icml-1994,littman-icml-2001,stimpson-goodrich-icml-2003,conitzer-sandholm-icml-2003,crandall-goodrich-icml-2005,powers-shoham-nips-2004,powers-shoham-ijcai-2005} and play matrix games (MG). Their performances have been assessed on well-chosen repeated MG (RMG). However, few publications systematically compare these various algorithms on a general and common test-bed RMG \cite{zawadzki-2005,airiau-2007,bouzy-metivier-icml-2010}.

Another stream of researches concerns the online prediction problem using expert advice. In this context, an hedging algorithm is made up with a top-level algorithm and a pool of expert algorithms. To predict the class of an instance, the top-level algorithm chooses an expert algorithm that predicts the class of the instance \cite{littlestone-warmuth-ic-1994,vovk-jcss-1998,freund-schapire-geb-1999,farias-megiddo-jacm-2006,chang-kaelbling-icml-2005}. 

The motivation of this paper is to merge the two contexts of RMG and hedging algorithms, and to experimentally demonstrate that, on RMG, well-chosen hedging algorithms are better on average than the previous MAL algorithms. This result is new and significant, and surpasses the previous ones \cite{zawadzki-2005,airiau-2007,bouzy-metivier-icml-2010}.

The paper is structured as follows. Section \ref{sectionRelatedWork} presents the settings and previous work addressing MAL assessment on RMG \cite{zawadzki-2005,airiau-2007,bouzy-metivier-icml-2010}. Section \ref{sectionHedgingAlgo} develops the interest of hedging algorithms. Section \ref{sectionExample} presents an example forseeing why hedging algorithms could be good at RMG, and launches the first significant experiment. Section \ref{sectionExperiment1} takes all previous MAL algorithms, and some well-chosen hedging algorithms, and experimentally demonstrates the superiority of specific hedging algorithms. Before the conclusion, section \ref{sectionTree} discusses the performance of hedging algorithms with several levels.

\section{ Previous work and settings } \label{sectionRelatedWork}

In this section, we describe the settings of our experiments, and we list the basic players selected.

\subsection{Settings} 

A two-player MG is a strategic game with two payoff matrix wherein each row corresponds to one elementary action of one player, and each column corresponds to one elementary action of the other player. Each player has its own matrix. For one player, the matrix entry at the intersection of a row and a column gives the outcome obtained by this player when this row and this column are chosen. The two players simultaneously choose their elementary action, and they obtain the return corresponding to their joint action. 

The evaluation framework used is the same as in \cite{bouzy-metivier-icml-2010}. There are two ways to evaluate algorithms: with or without elimination. First, the version without elimination is based on several experiments. One experiment consists in drawing $N$ matrix games at random with return values in the interval $[-9, +9]$, and executes $N$ RMG tournaments, one for each MG generated. In a RMG tournament, each player meets all players in a RMG match. A RMG match is a game consisting in playing the same MG repetitively, and cumulating the outcomes obtained at each repetition. The default number of repetitions (or steps) is $100,000$. In a RMG match, the players try to maximize their cumulative return. At the end of a RMG tournament, a player has a cumulative return over all its matches. In an experiment, the output of an algorithm is its cumulative return over repetitions and matrix games played. The output of an experiment is a ranking. 

The relative rank of two given algorithms, $A$ and $B$, depends on the presence, or absence, of other players. For instance, if algorithm $A$ performs well against algorithm $C$ by gathering 100 points against it, and if algorithm $B$ plays poorly against algorithm $C$ by gathering no point against it, then, an all-against-all tournament with $A$, $B$ and $C$, ranks $A$ 100 points better than a tournament without $C$. Consequently, $A$ can be ranked before or after $B$ depending on the presence or absence of $C$. Therefore, to lower this effect, the elimination principle of \cite{bouzy-metivier-icml-2010} is used. It consists in launching an experiment with a large number of players and removing the worst player when his cumulative return is significantly behind the before-last player or when the global ranking has not been changed during a long time. When a player is removed, all the returns obtained against him from the beginning of the experiment are suppressed from the cumulative returns of other agents. This removing process is applied at the end of each RMG and the experiment is continued until only one player remains.

\subsection{Basic Players} 

Three papers help us to select the pool of basic players. First, Zawadzki \cite{zawadzki-2005} compares several algorithms including Nash equilibrium oriented algorithms \cite{conitzer-sandholm-icml-2003,littman-icml-1994,singh-kearns-mansour-uai-2000,banerjee-peng-aaai-2004}, Q-learning ($Q$) \cite{watkins-dayan-1992}, and portfolio algorithms \cite{powers-shoham-nips-2004,powers-shoham-ijcai-2005} on instances of MG generated with GAMUT \cite{nudelman-al-amaas-2004}, concluding on the superiority of $Q$. This result confirms that, against various and unknown players, using the concept of Nash equilibrium is worse than being a best response (BR) to other players. Second, Airiau \cite{airiau-2007} compares several algorithms, including some learning ones, and concludes on the superiority of Fictitious Play ($F$) \cite{brown-51}. Third, Bouzy \cite{bouzy-metivier-icml-2010} provides new evaluations taking other learners such as $UCB$ ($U$) \cite{auer-ml2002}, $Exp3$ \cite{auer-siam2002}, $S$ \cite{stimpson-goodrich-icml-2003}, $M3$ \cite{crandall-goodrich-icml-2005}, $J$ \cite{robinson-am-1951} into account, and concludes on the superiority of $M3$ and $U$ on general-sum games, and on the superiority of $Exp3$ on zero-sum games and on cooperative games. Moreover, Bouzy \cite{bouzy-metivier-icml-2010} experimentally demonstrates that two features, the state heuristic and the fixed-window heuristic, are positive enhancements to algorithms that do not own them. The window heuristic ($w$) consists in updating the mean values with a constant rate $\alpha=0.01$. The state heuristic ($s$ in the following) \cite{sandholm-crites-biosystems-1996} consists in associating a state to the last joint action. In this paper, when a player $P$ is enhanced with the two heuristics, its name becomes $Pw+s$.

The three papers above lead us to consider $Q$, $F$, $J$, $U$, $Exp3$, $M3$, $S$, $Uw+s$, $Qw+s$, and $Jw+s$ as promising candidates for the current work. Moreover, we also add the greedy player ($G$), the random player $Rand$ ($R$), $MinMax$, and $Bully$ ($B$) \cite{littman-stone-atal-2001}. $B$ chooses the action that maximizes its own return assuming its opponent is a BR learner. Assuming its opponent keeps its action unchanged, $G$ chooses the action with the best return. $R$ plays randomly according to the uniform probability.

Each algorithm may have specific parameters whose values are tuned experimentally. For every action, taken or not, $J$ saves the cumulative return, and selects the action with the best cumulative return. $MinMax$ guarantees reaching at least the minmax value on every game ($Determined$ in \cite{zawadzki-2005}, or $Nash$ in \cite{airiau-2007}). $Q$ has two parameters: $\gamma=0.95$ and $\alpha=1/t$, $t$ being the timestep of the current game. The $\epsilon$-greedy method is used with $\epsilon=1/\sqrt{t/n_a}$, $n_a$ being the number of actions. $S$ has two parameters: $\alpha=12$ and $\lambda=0.99$. $M3$ has three parameters: $\gamma=0.95$, $\alpha=0.1$, and $\lambda=0.01$. $U$ selects the action $j$ with the highest value $U(j) = \bar{x_j} + \sqrt{C\ln{(t)}/n_j}$, where $\bar{x_j}$ is the average return obtained when action $j$ is played, $t$ is the timestep and $n_j$ is the number of times action $j$ has been played. $C$ is a parameter set to $100$. $Exp3$ has one parameter: $\gamma=0.001$. 

Each algorithm has its own information requirements: the whole matrix ($MinMax$, $G$ and $B$), the opponent actions and the actual return ($F$), all the returns corresponding to all actions played or not ($J$), the actual return only ($Q$, $S$, $U$, $Exp3$), or nothing ($R$). $M3$ needs the actual return but also the minmax value, and consequently the matrix. All the experiments let all the players have their requirements satisfied.

\section{Hedging algorithms} \label{sectionHedgingAlgo}

This section underlines work related to hedging algorithms, explains what an algorithm game is, and finally gives our definition of hedging algorithms.

\subsection{Related work} 

In the binary online prediction problem, an instance is presented, and the solver must predict the class (0 or 1) of the instance. The weighted majority algorithm \cite{littlestone-warmuth-ic-1994} asks its experts for a binary prediction. Each expert has a weight. Each class ($0$ or $1$) has a weight which is the sum of the weight of the experts agreeing with the class. The master algorithm takes the class with the highest weight as its prediction. Then the expert weights are decreased when they made a mistake. The weighted majority algorithm is proven to be robust in presence of errors in the data. The binary prediction problem is not similar to the RMG problem. However, the idea of using experts remains interesting to try with the RMG problem. The aggregating algorithm \cite{vovk-jcss-1998} can be a solution nearer to the RMG problem requirements, since experts are used, and there is a loss function giving a numerical return to a prediction. The multiplicative weight algorithm \cite{freund-schapire-geb-1999} is a hedging algorithm designed to play matrix games. To this extent, it fits with our settings. However, since the multiplicative weight algorithm is an ancestor of $Exp3$, inferior to $Exp3$ on RMG, and since $Exp3$ is already part of our players, we do not consider the multiplicative weight algorithm. The Exploration-Exploitation Experts method \cite{farias-megiddo-jacm-2006} and the Hierarchical Hedging algorithm \cite{chang-kaelbling-icml-2005} address decision problems using experts explicitly. They are used on the prisoner dilemma matrix game. They define the concept of master algorithm, or top-level algorithm: the algorithm that must decide which expert will choose the basic action. The expert is chosen for a given period in which it chooses a basic action.

\subsection{The algorithm game} 

In matrix games, having a top-level algorithm using a pool of basic algorithms may mean playing a super-game \cite{chang-kaelbling-icml-2005}, or an algorithm game \cite{zawadzki-2005}. The algorithm game can be defined by assuming that two top-level algorithms are playing a common matrix game in which they have to choose an algorithm rather than an action. Each row and each column of the algorithm game matrix corresponds to an expert algorithm. After an all-against-all tournament between a pool of basic algorithms, the mean results of the two matches between expert A and expert B may be written in the cell situated at row A and column B. This yields a matrix of a game, called the super-game or the algorithm game. 

\subsection{Definition} 

In the current paper, we call $Hedge(top=T, A, B, ...)$ the algorithm which is made up with $T$ as top-level algorithm (or master algorithm) and with $A$, $B$, etc, as basic algorithms, or expert algorithms. When $Hedge(top=T, A, B, ...)$ has to choose an elementary action, $top$ chooses a basic algorithm in the set $\{A, B, ...\}$. Then, the chosen basic algorithm chooses an elementary action. When the context is not ambiguous, we use the symbol $H$ for simplicity. An important feature of a hedging algorithm is updating. In our study, we assume that, after the reward is received, the master algorithm and the chosen basic algorithm update their values, and not the basic algorithms which were not chosen by the master. To simplify the study, the master algorithm chooses its basic algorithm for one step only. This simplification is not an obstacle to achieve very good results.

\section{Example} \label{sectionExample}

In this section, we give an example showing the relevance of hedging algorithms on RMG. Let us consider an experiment without elimination between three basic algorithms $U$, $G$, and $S$ on a set of $1,000$ random RMG. Table \ref{tab-example-base} shows the result of the two-by-two confrontations. Each cell contains the average return obtained by the two players, row and column, after the confrontation. For each player playing rows, column $rt$ contains the total of the average returns obtained by the player. Similarly, for each player playing columns, $ct$ contains the column total. Column $T$ contains $rt+ct$ for each player. The higher the $T$, the better the player. We see that $U$ is the best player, before $S$ and $G$.

\begin{table}[ht]
\caption{Basic experiment results.}
\begin{center}
\begin{tabular}{|l|l|l|l|l|l|}
   \hline       &    $U$       &     $G$      & $S$             &   rt    &  T      \\ 
   \hline  $U$  &  4.45  4.42  & 4.11  4.48   & 5.77      4.33  &  14.33  &  28.67  \\
   \hline  $G$  &  4.53  4.07  & 3.12  3.04   & 4.59      4.35  &  12.24  &  24.53  \\
   \hline  $S$  &  4.15  5.86  & 4.35  4.77   & 4.85      4.99  &  13.34  &  27.01  \\
   \hline  ct   &       14.35  &       12.29  &          13.67  &         &         \\
   \hline   
\end{tabular}
\label{tab-example-base}
\end{center}
\end{table}

Table \ref{tab-example-base} can be considered as the matrix of an algorithm game, or super-game, in which the players have to choose an algorithm rather than an action. The question is to choose the algorithm that will best play this experiment. What would be the performances of a ``perfect'' player $H$ that could choose the basic algorithm maximizing its average return when playing against a given opponent? 

For instance, if $H$ plays column against $S$ playing rows, his choice would be $U$ to obtain the $5.86$ return. Playing columns against $G$ playing rows, his choice would be $S$ to obtain $4.35$. Playing rows against $S$ playing columns, his choice would be $U$ to obtain $5.77$, and so on. 

The answer is given on table \ref{tab-example-expectation} which shows the predictions of the average return obtained by $H$. First, the cells of the first three rows and columns of table \ref{tab-example-expectation} are identical to the corresponding cells of table \ref{tab-example-base}. Second, for $l$ in $\{U, G, S\}$, the two numbers in column $H$ and row $l$ are equal to the two numbers of the cell of row $l$ and the column maximizing the second number over the columns. For instance, cell $(U, H)$ contains $4.11$ and $4.48$ because $4.48$ is the maximum of the second numbers of row $U$ over its columns. Similarly, for $c$ in $\{U, G, S\}$, the two numbers in row $H$ and column $c$ are equal to the two numbers of the cell of column $c$ and the row maximizing the first number over the rows. For instance, cell $(H, U)$ contains $4.53$ and $4.07$ because $4.53$ is the maximum of the first numbers of column $U$ over its rows. Finally, which values can be expected for $H$ in self-play? If $H$ playing rows, and $H$ playing columns, adapt well to each other, they may find a joint action in which the sum of the two average returns is the highest. $(U, S)$ is the best joint action ($10.10$). Table \ref{tab-example-expectation} shows that $H$ obtains the best total $T$ ($39.44$), and surpasses $U$ ($36.86$), $S$ ($35.49$) and $G$ ($33.89$). 

\begin{table}[ht]
\caption{Expected results with a perfect player $H$.}
\begin{center}
%{\footnotesize%{\small%{\tiny
{\scriptsize
\begin{tabular}{|l|l|l|l|l|l|l|}
\hline       &    $U$       &      $G$     &   $S$       &   $H$        &   rt   &  T      \\ 
\hline  $U$  &  4.45  4.42  & 4.11  4.48   & 5.77  4.33  &  4.11  4.48  & 18.44  &  36.86  \\
\hline  $G$  &  4.53  4.07  & 3.12  3.04   & 4.59  4.35  &  4.59  4.35  & 16.83  &  33.89  \\
\hline  $S$  &  4.15  5.86  & 4.35  4.77   & 4.85  4.99  &  4.15  5.86  & 17.49  &  35.49  \\
\hline  $H$  &  4.53  4.07  & 4.35  4.77   & 5.77  4.33  &  5.77  4.33  & 20.42  &  39.44  \\
\hline   ct  &       18.42  &      17.06   &      18.00  &       19.02  &        &         \\
\hline   
\end{tabular}
\label{tab-example-expectation}
}
\end{center}
\end{table}

Now, let us see whether an adequate hedging algorithm can actually achieve the expected results of table \ref{tab-example-expectation}. The simplest pool of basic algorithms is $\{U, G, S\}$, and we have the choice for $top$. Since $S$ is the simplest algorithm of our set, we take $top=S$. Of course, other choices are possible. Let us take $H = HSUGS = Hedge(top=S, U, G, S)$, and launch the experiment. The set of $1,000$ MG used here is the same set of MG used for table \ref{tab-example-base}. Table \ref{tab-example-results} provides us with the results of an experiment between the basic algorithms and $H = Hedge(top=S, U, G, S)$.

\begin{table}[ht]
\caption{The results of an experiment between the basic algorithms and $H = HSUGS = Hedge(top=S, U, G, S)$.}
\begin{center}
{\scriptsize
\begin{tabular}{|l|l|l|l|l|l|l|}
\hline     &   $U$       & $G$        &  $S$       &  $H$        &   rt    &   T    \\ 
\hline $U$ & 4.42  4.39  & 4.11  4.47 & 5.89  4.26 & 5.30  4.75  &  19.73  &  39.30 \\
\hline $G$ & 4.51  4.07  & 3.18  3.07 & 4.70  4.38 & 1.12  5.77  &  13.51  &  27.21 \\
\hline $S$ & 4.12  5.84  & 4.36  4.68 & 4.78  4.93 & 4.36  5.37  &  17.62  &  35.57 \\
\hline $H$ & 4.68  5.26  & 5.61  1.49 & 5.29  4.38 & 4.92  4.93  &  20.49  &  41.32 \\
\hline  ct &      19.57  &      13.71 &      17.95 &      20.83  &         &        \\
\hline   
\end{tabular}
\label{tab-example-results}
}
\end{center}
\end{table}

Beforehand, it is worth noticing that, because they refer to the same confrontations, the values of cell $(x, y)$ in table \ref{tab-example-results} and in table \ref{tab-example-base} (for $x$ and $y$ in $\{U, G, S\}$) can be expected to be the same. However, the values are not identical. The explanation does not correspond to the random creation of the MG since we used the same sets in the two experiments, but to the order of confrontations used in the two experiments. Because some players use randomness to make their choices, the order of the confrontations has an impact on the choice of actions, which explains the differences. The differences observed between the two tables correspond to the standard deviation of our experiment. The maximal difference is $0.12$, and the average difference is $0.02$.

First, the global result is very good: $H$ obtains a cumulative return significantly higher than the one of the basic algorithms. The difference observed between $H$ and $U$ average returns is about $2.0$, which confirms the expected difference of $2.6$. Table \ref{tab-example-expectation} and table \ref{tab-example-results} show that $H$ does better than expected against $G$ (plus $2.7$ on average), and $U$ (plus $0.4$ on average), does worse than expected against $S$ (minus $1.0$), and does worse than expected in self-play (minus $0.2$). 

Second, $H$ playing columns against $S$ or $H$, obtains an average reward ($5.37$ or $4.93$) inferior to the average reward obtained by $U$ against them ($5.84$ or $5.26$). Therefore, in practice, $H$ does not dominate every basic player on every confrontation. There is still room for improvement, either by improving $H$ itself, or by repeating the process of adding a new hedging algorithm using $U$, $G$, $S$, $H$ (see subsection \ref{subsectionTwoLevels}).

To sum up, in practice, the hedging algorithm $H = HSUGS = Hedge(top=S, U, G, S)$ performs better on average than $U$, $G$, $S$, which was expected by our considerations if the master algorithm can choose the proper expert quickly. On this example, we conclude that $H$ adapts very quickly to its opponent by choosing the proper basic algorithm.

\section{Experiments} \label{sectionExperiment1}

In this section, we experimentally assess some well-chosen hedging MAL algorithms $H$ against $U$, $M3$, $S$, $Q$, $J$, and $F$. The master algorithm and the pool of experts is made up with these classical MAL algorithms too. $H = H(top, x, y)$, with $top$, $x$ and $y$ taken in $\{F, U, S, J, Q, M3\}$. After a preliminary set of experiments, we observed that $top$ cannot be $F$, $U$, $M3$, $Q$, or $J$, whatever the basic algorithms $x$ and $y$. The unique possible solution was $top=S$. Table \ref{tab5} sums up the results by giving an evaluation to each $H$. The evaluation can be $--$, $=$, $+$, or $++$. $--$ means that $H$ is ranked in the middle of the league of players $\{ H, U, S, J, Q, M3, F, B, Minmax, R\}$. $=$ means $H$ is a runner up of the league (but not ahead). $+$ means that $H$ is ranked first. $++$ means that $H$ is ranked first with a significant margin before the runner up.

%\begin{table}[ht]
%  \centering
%  \subfloat[Subtable 1 list of tables text][Subtable 1 caption]{
%    \begin{tabular}{l|ccc}
%      & 1 & 2 & 3\\
%      \hline
%      1 & A & B & C\\
%      2 & D & E & F\\
%    \end{tabular}}
%  \qquad
%  \subfloat[Subtable 2 list of tables text][Subtable 2 caption]{
%    \begin{tabular}{l|ccc}
%      & 1 & 2 & 3\\
%      \hline
%      1 & A & B & C\\
%      2 & D & E & F\\
%    \end{tabular}}
%  \caption{This is a table containing several subtables.}
%\end{table}
	
\begin{table}[ht]
  \centering
  \subfloat[][The results of several attempts of $H = H(top, x, y)$.]{
    \begin{tabular}{|l|l|l|}
      \hline     $H$              &  evaluation \\ 
      \hline     $H(S, U, U)$     &   --        \\ 
      \hline     $H(S, U, S)$     &   --        \\ 
      \hline     $H(S, J, J)$     &   --        \\ 
      \hline     $H(S, J, S)$     &   =         \\ 
      \hline     $H(S, S, S)$     &   =         \\ 
      \hline     $H(S, Q, S)$     &   =         \\ 
      \hline     $H(S, Q, Q)$     &   =         \\ 
      \hline     $H(S, M3, S)$    &   +         \\ 
      \hline     $H(S, M3, M3)$   &   +         \\ 
      \hline     $H(S, M3, J)$    &   ++        \\ 
      \hline     $H(S, U, M3)$    &   ++        \\ 
      \hline   
    \end{tabular}
\label{tab5}
  }
  \qquad
  \subfloat[][Experiment with $HSUM = H(S, U, M3)$ in the league.]{
    \begin{tabular}{|l|l|l|}
      \hline     &                  &  Av. return  \\
      \hline  1  &     $HSUM$       &   4.88       \\ 
      \hline  2  &     $M3$         &   4.74       \\ 
      \hline  3  &     $U$          &   4.60       \\ 
      \hline  4  &     $J$          &   4.15       \\ 
      \hline  5  &     $S$          &   4.17       \\ 
      \hline  6  &     $B$          &   4.47       \\ 
      \hline  7  &     $Q$          &   4.49       \\ 
      \hline  8  &     $MinMax$     &   4.64       \\ 
      \hline  9  &     $R$          &  -2.78       \\ 
      \hline
    \end{tabular}
\label{tab6}
  }
  \qquad
  \subfloat[][Experiment with $Uw+s$, $M3$, $Jw+s$, $S$, $Qw+s$, and $H = Hedge(top=S, U, M3)$.]{
    \begin{tabular}{|l|l|l|}
      \hline     &     Player       & Av. return  \\ 
      \hline  1  &     $HSUM$       &   5.03      \\ 
      \hline  2  &     $Uw+s$       &   4.67      \\ 
      \hline  3  &     $M3$         &   4.57      \\ 
      \hline  4  &     $Jw+s$       &   4.47      \\ 
      \hline  5  &     $S$          &   3.80      \\ 
      \hline  6  &     $Qw+s$       &   3.58      \\ 
      \hline   
    \end{tabular}
\label{tab4}
  }
\caption{}
\end{table}

$H(S, S, S)$ is ranked near $M3$ and $U$, and before $S$. $Q$ may be worth considering as a basic algorithm of $H$, but the appropriate basic algorithms are $U$, $J$, and $M3$. $HSUM = H(S, U, M3)$ and $H(S, J, M3)$ are the best combinations found. When launching $HSUM$ in a league, it ranked first, as shown by table \ref{tab6}.

In order to explicitly prove the superiority of the best hedging algorithms over the best MAL algorithms, we assess $HSUM$ against $Uw+s$, $Qw+s$, $Jw+s$, $S$, and $M3$.

Table \ref{tab4} shows that $HSUM$ outperforms $Uw+s$, $Qw+s$, $Jw+s$, $S$, and $M3$. It is worth noticing here that $HSUM$ uses $U$, and not $Uw+s$ as a basic algorithm. This shows that embedding $M3$ and $U$ with $top=S$ into $HSUM$ is a more efficient solution than adding the two features $s$ and $w$ into $U$. Besides, we observed an amazing result: $HSSS = Hedge(top=S, x=S, y=S)$ is significantly better than $S$. Hedging algorithms offer the possibility to incorporate many opponent models and strategies into its framework. When the opponents change their strategies during play, hedging algorithms easily adapt to the new strategies. When playing against an opponent whose model is not incorporated into the framework, the various strategies already incorporated enable hedging algorithms to perform not too badly.

\section{Several levels} \label{sectionTree}

This section raises the question of using several levels of hedging algorithms, i.e. using hedging algorithms as basic algorithms of an hedging algorithm. 

\subsection{Two levels} \label{subsectionTwoLevels}

What is the effect of adding a new perfect hedge player $HH = Hedge(top, U, G, S, H)$ in table  \ref{tab-example-expectation} that would use perfectly $U$, $G$, $S$, and $H$ as basic algorithms? Trivially, since $HH$ would maximize actions already maximized by $H$, $HH$ would obtain the same returns than $H$. Therefore, theoretically two levels are of no interest. However, table \ref{tab-example-results} shows that in pratice, although $H$ has the best cumulative return, $H$ does not obtain the best result against some opponents. Therefore, table \ref{tab-example-results} could be extended by adding another hedging algorithm that would use its algorithms as basic algorithms. In practice, two level hedging algorithms have to be assessed.

Table \ref{tab-example-twolevels-results} shows the actual results of $HH = Hedge(top=S, HSUGS, U, G, S)$ against $HSUGS$ ($H$), $U$, $G$, $S$. $HH$ obtains the best cumulative return ($51.7$), significantly above the second best cumulative return ($51.0$), which confirms the significance of two levels in practice. However, against specific players, $HH$ does not dominate the other players' results. For instance, against $S$, $H$ or $HH$ playing rows respectively, $HH$ playing columns obtains an average return (respectively $5.5$, $5.0$, $4.9$) inferior to the average return obtained by $U$ (respectively $5.9$, $5.3$, $5.3$).

\begin{table}[ht]
\caption{The results of an experiment between $U$, $G$, $S$, $H$, and $HH$.}
\begin{center}
%{\footnotesize%{\small%{\tiny
{\scriptsize
\begin{tabular}{|l|l|l|l|l|l|l|}
\hline     &    $U$   &   $G$    &   $S$    &   $H$    &   $HH$   &   T  \\ 
\hline $U$ & 4.5  4.4 & 4.1  4.5 & 5.8  4.3 & 5.3  4.8 & 5.3  4.8 & 50.0 \\
\hline $G$ & 4.5  4.1 & 3.1  3.1 & 4.6  4.4 & 1.3  5.7 & 1.0  6.0 & 29.5 \\
\hline $S$ & 4.1  5.9 & 4.4  4.6 & 4.8  5.0 & 4.4  5.4 & 4.2  5.5 & 44.4 \\
\hline $H$ & 4.7  5.3 & 5.6  1.5 & 5.2  4.3 & 4.9  5.0 & 4.7  5.0 & 51.0 \\
\hline $HH$& 4.6  5.3 & 5.8  1.3 & 5.3  4.4 & 4.9  4.9 & 4.8  4.9 & 51.7 \\
\hline   
\end{tabular}
\label{tab-example-twolevels-results}
}
\end{center}
\end{table}

As in section \ref{sectionExperiment1}, we tried several combinations of two level hedging algorithms. A lot of them do not work well and few work. Let $HHUMM = Hedge(top=S, HSUM, M3)$, $HHUMM$ is a two-level hedging algorithm using two basic algorithms, $HSUM$ and $M3$. Let us compare $HHUMM$ with $HSUM$, $M3$, $U$, $S$. Table \ref{tab-twolevels-league} shows the results of an experiment with elimination, with the average return at the time of elimination.

\begin{table}[ht]
  \centering
  \subfloat[][Experiment with two levels in the league.]{
    \begin{tabular}{|l|l|l|}
      \hline       & Player     &   Av. ret. \\ 
      \hline   1   & $HHUMM$    &   5.044   \\ 
      \hline   2   & $HSUM$     &   4.956   \\ 
      \hline   3   & $M3$       &   4.46    \\ 
      \hline   4   & $U$        &   4.444   \\ 
      \hline   5   & $S$        &   3.788   \\ 
      \hline
    \end{tabular}
    \label{tab-twolevels-league}
  }
  \qquad
  \subfloat[][Experiment with $HSUM$, $Uw+s$, $M3$, $Jw+s$, $S$, $Qw+s$, and $HHUMM$.]{
    \begin{tabular}{|l|l|l|}
      \hline     &     Player       & Av. ret.  \\ 
      \hline  1  &     $HHUMM$      &   4.94      \\ 
      \hline  2  &     $HSUM$       &   4.88      \\ 
      \hline  3  &     $Uw+s$       &   4.72      \\ 
      \hline  4  &     $M3$         &   4.56      \\ 
      \hline  5  &     $Jw+s$       &   3.46      \\ 
      \hline  6  &     $S$          &   3.20      \\ 
      \hline  7  &     $Qw+s$       &   3.21      \\ 
      \hline   
    \end{tabular}
    \label{tab-twolevels-prove}
  }
  \qquad
  \subfloat[][Experiment with three levels in the league.]{
    \begin{tabular}{|l|l|l|}
      \hline       & Player       &  Av. ret.  \\ 
      \hline   1  & $HHUMM$       &  5.078         \\ 
      \hline   2  & $HHHUMM$      &  5.053         \\ 
      \hline   3  & $HSUM$        &  5.131         \\ 
      \hline   4  & $M3$          &  4.77          \\ 
      \hline   5  & $U$           &  4.583         \\ 
      \hline   6  & $S$           &  4.075         \\ 
      \hline
    \end{tabular}
    \label{tab-threelevels-league}
  }
  \caption{}
\end{table}

In this experiment, $HHUMM$ is significantly the best player ahead of $HSUM$, the one-level champion. We tried other two level hedging algorithms combinations such as $HHUMMM = Hedge(top=S, HSUM, M3, M3)$. $HHUMM$ remains the best two-level algorithm. Finally, in order to prove the superiority of the best two level hedging algorithms over the best MAL algorithms, we assessed $HHUMM$ against $HSUM$, $Uw+s$, $Qw+s$, $Jw+s$, $S$, and $M3$ in an experiment with the elimination mechanism. Table \ref{tab-twolevels-prove} shows that $HHUMM$ outperforms $HSUM$, $Uw+s$, $Qw+s$, $Jw+s$, $S$, and $M3$.

\subsection{Three levels}

Since two levels work better than one level in practice, let us see what happens for three levels. Let us define three level hedging algorithms and assess their playing level against one-or-two-level hedging algorithms. Let $HHHUMM = Hedge(top=S, HHUMM, M3)$. $HHHUMM$ is a three-level hedging algorithm using two basic algorithms, $HHUMM$ and $M3$. Let us compare $HHHUMM$ with $HHUMM$, $HSUM$, $M3$, $U$, $S$. Table \ref{tab-threelevels-league} shows the results of an experiment with elimination, with the average return at the time of elimination.

In this experiment, $HHHUMM$ remains below $HHUMM$ the two-level champion. We tried other three level hedging algorithm combinations without success. Furthermore, we extended the number of repetitions to see whether the three-level hedging algorithms need more repetitions to reach stability, but even with $1,000,000$ repetitions instead of $100,000$, the results remain the same. So we observe that, in practice, three levels are not better than two levels.

An explanation of these practical observations about levels could remain on the number of internal parameters of the algorithms that adapt accordingly to the returns received during play or not. The higher the number of levels, the higher the number of parameters, which would explain the increase between one level and two levels. But the higher the number of levels, the smaller the precision of parameter values, which would explain the decrease between two levels and three levels. We did not launch experiments with more than three levels.

\section{Conclusion} \label{sectionConclusion}

This paper extends the results of \cite{zawadzki-2005,airiau-2007,bouzy-metivier-icml-2010} on existing MAL algorithms playing RMG. Given the promising expert algorithm litterature, this paper experimentally studied hedging algorithms on RMG. First, the results highlight that specific hedging algorithms work significantly better than previous MAL algorithms \cite{bouzy-metivier-icml-2010} on RMG. This result is new and significant in the multi-agent learning field. Second, $S$ is our unique solution for a master algorithm. Furthermore, although two level hedging algorithms are comparable with one level hedging algorithms in theory, we actually observed that two levels perform better on average than one level. We also observed experimentally that three levels are not better than two levels. Those very good results can be due to the capacity of hedging algorithms to incorporate many opponent models, and various strategies. Even when the opponent changes its strategy, hedging algorithms can adapt to the new opponent strategy.

This experimental research can be investigated in several directions. First, exploring the hedging algorithm space more exhaustively with genetic algorithms to find better combinations of basic algorithms remains to be done. For instance, we explored solutions with two or three basic expert algorithms, but the principle of hedging has no limit on the number of basic algorithms. We also explored solutions with $S$ as master algorithm. Solutions with other master algorithms probably exist. In the positive case, we have to find out which ones, and in the negative case, we have to get a better understanding about why $S$ works so well as top algorithm. We also limited our study to two or three levels of hedging algorithms, but solutions with more levels may exist too. It is possible that building a tree of experts could work well, although it is impossible to be optimal with respect to all possible opponents \cite{nachbar-et-1996}. Second, seeing whether the results of our hedging algorithms can be extended to other frameworks, domains, or stochastic games, would be informative. Third, developing an appropriate theory explaining the results obtained practically becomes crucial to go further. Fourth, observing the result of a master algorithm following its selected basic algorithm for several steps \cite{chang-kaelbling-icml-2005}, instead of one step, would extend our work.

\section{Acknowledgments}
This work has been supported by French National Research Agency (ANR) through COSINUS program (project EXPLO-RA number ANR-08-COSI-004).

%\bibliography{mg}
% ---- Bibliography ----

%\clearpage
%\addtocmark[2]{Author Index}
%\renewcommand{\indexname}{Author Index}
%\printindex
%\clearpage
%\addtocmark[2]{Subject Index} % additional numbered TOC entry
%\markboth{Subject Index}{Subject Index}
%\renewcommand{\indexname}{Subject Index}
%\input{subjidx.ind}

\end{document}